\documentclass{article}

    \usepackage[preprint]{neurips_2019}

\usepackage[utf8]{inputenc} 
\usepackage[T1]{fontenc}    
\usepackage{hyperref}       
\usepackage{url}            
\usepackage{booktabs}       
\usepackage{amsfonts}       
\usepackage{nicefrac}       
\usepackage{microtype}      
\usepackage{numprint}
\usepackage{float}
\usepackage{breakcites}
\usepackage[all]{hypcap}

\usepackage[dvipsnames]{xcolor}
\usepackage[normalem]{ulem}

\usepackage{graphicx, multirow}

\newif{\ifhidecomments}
\usepackage{graphicx}

\title{Reproducibility Study: \emph{Comparing Rewinding\\and Fine-tuning in Neural Network Pruning}}

\author{%
 Szymon Mikler \\
  Uniwersytet Wrocławski\\
  \texttt{sjmikler@gmail.com} \\
}

\begin{document}

\maketitle

\section*{\centering Reproduction Summary}


\subsection*{Scope of Reproducibility}
    
We are reproducing \emph{Comparing Rewinding and Fine-tuning in Neural Networks}, by \cite{Renda}.
In this work the authors compare three different approaches to retraining neural networks after pruning: 1) fine-tuning, 2) rewinding weights as in \cite{Frankle} and 3) a new, original method involving learning rate rewinding, building upon \cite{Frankle}. We reproduce the results of all three approaches, but we focus on verifying their approach, learning rate rewinding, since it is newly proposed and is described as a universal alternative to other methods.

We used CIFAR10 for most reproductions along with additional experiments on the larger CIFAR100, which extends the results originally provided by the authors. We have also extended the list of tested network architectures to include Wide ResNets (\cite{wrn}). The new experiments led us to discover the limitations of learning rate rewinding which can worsen pruning results on large architectures.

\subsection*{Methodology}

We implemented the code ourselves in Python with TensorFlow 2, basing our implementation of the paper alone and without consulting the source code provided by the authors. We ran two sets of experiments. In the reproduction set, we have striven to exactly reproduce the experimental conditions of \cite{Renda}. We have also conducted additional experiments, which use other network architectures, effectively showing results previously unreported by the authors. We did not cover all originally reported experiments -- we covered as many as needed to state the validity of claims. We used Google Cloud resources and a local machine with 2x RTX 3080 GPUs.

\subsection*{Results}

We were able to reproduce the exact results reported by the authors in all originally reported scenarios. However, extended results on larger Wide Residual Networks have demonstrated the limitations of the newly proposed learning rate rewinding -- we observed a previously unreported accuracy degradation for low sparsity ranges. Nevertheless, the general conclusion of the paper still holds and was indeed reproduced.

\subsection*{What was easy}

Re-implementation of the pruning and retraining methods was technically easy, as it is based on a popular and simple pruning criterion -- magnitude pruning. Original work was descriptive enough to reproduce the results with satisfying results without consulting the code.

\subsection*{What was difficult}

Not every design choice was mentioned in the paper, thus reproducing the exact results was rather difficult and required a meticulous choice of hyper-parameters. Experiments on ImageNet and WMT16 datasets were time consuming and required extensive resources, thus we did not verify them.


\subsection*{Communication with original authors}

We did not consult the original authors, as there was no need to.

\newpage
\renewcommand{\arraystretch}{1.5}

\section{Introduction}

Neural network pruning is an algorithm leading to decrease the size of a network, usually by removing its connections or setting their weights to 0. 
This procedure generally allows obtaining smaller and more efficient models.
It often turns out that these smaller networks are as accurate as their bigger counterparts or the accuracy loss is negligible.
A common way to obtain such high quality sparse network is to prune it after the training has finished (\cite{rethinking}, \cite{Frankle}).
Networks that have already converged are easier to prune than randomly initialized networks (\cite{rethinking}, \cite{snip}).
After pruning, more training is usually required to restore the lost accuracy.
Although there are a few ways to retrain the network, finetuning might be the easiest and most often chosen by researchers and practitioners (\cite{rethinking}, \cite{Renda}).

Lottery Ticket Hypothesis from \cite{Frankle} formulates a hypothesis that for every dense neural network, there exists a smaller subnetwork that matches or exceeds results of the original. The algorithm originally used to obtain examples of such networks is iterative magnitude pruning with weight rewinding, and it is one of the methods of retraining after pruning compared in this work.


\section{Scope of reproducibility}
\label{sec:claims}



\cite{Renda} formulated the following claims:
\begin{itemize}
    \item [Claim 1:] Widely used method of training after pruning: finetuning yields worse results than rewinding based methods (supported by figures \ref{fig:resnet20-1}, \ref{fig:resnet56}, \ref{fig:resnet20-2}, \ref{fig:resnet56-2} and Table \ref{tab:resnet110})
    \item [Claim 2:] Newly introduced learning rate rewinding works as good or better as weight rewinding in all scenarios (supported by figures \ref{fig:resnet20-1}, \ref{fig:resnet56}, \ref{fig:resnet20-2}, \ref{fig:resnet56-2} and Table \ref{tab:resnet110}, but not supported by Figure \ref{fig:wrn-1})
    \item [Claim 3:] Iterative pruning with learning rate rewinding matches state-of-the-art pruning methods \\(supported by figures \ref{fig:resnet20-1}, \ref{fig:resnet56}, \ref{fig:resnet20-2}, \ref{fig:resnet56-2} and Table \ref{tab:resnet110}, but not supported by Figure \ref{fig:wrn-1})
\end{itemize}



\section{Methodology}


We aimed to compare three retraining approaches: 1) finetuning, 2) weight rewinding and 3) learning rate rewinding. Our general strategy that repeated across all experiments was as follows:
\begin{enumerate}
    \item train a dense network to convergence,
    \item prune the network using magnitude criterion: remove weights with smallest L1 norm,
    \item retrain the network using selected retraining approach.
\end{enumerate}

In the case of structured pruning: in step 2, we removed structures (rows or convolutional channels) with the smallest average L1 norm (\cite{structured}), rather than removing separate connections. 

In the case of iterative pruning: the network in step 1 was not randomly initialized, but instead: weights from a model from a previous iterative pruning step were loaded as the starting point.

We trained all our networks using Stochastic Gradient Descent with Nesterov Momentum. The learning rate was decreased in a piecewise manner during the training, but momentum coefficient was constant and equal to $0.9$.

\subsection{Model descriptions}

In this report, we were focusing on an image recognition task using convolutional neural networks (\cite{lecun_convnet}). 
For most of our experiments, we chose to use identical architectures as \cite{Renda} to better validate their claims and double-check their results, rather than provide additional ones.
Therefore, most of the used networks are residual networks, which were originally proposed in \cite{resnet}.
Additionally, to verify the general usefulness of pruning and retraining methods proposed in \cite{Renda} we extend the list of tested network architectures to much larger wide residual networks from \cite{wrn}.

\subsubsection{Residual networks (ResNet)}

Just as \cite{Renda}, we chose to use the original version of ResNet as described in \cite{resnet} rather than the more widely used, improved version (with preactivated blocks) from \cite{resnetv2}. We created the models ourselves, using TensorFlow (\cite{tensorflow}) and Keras. We strove to replicate the exact architectures used by \cite{Renda} and \cite{resnet} and train them from scratch.

\begin{table}[H]
\small
\setlength{\tabcolsep}{10pt}
  \begin{center}
    \begin{tabular}{l|c|c|c|c}
      \specialrule{1pt}{2pt}{2pt}
    \textbf{Model} & \textbf{Trainable parameters} & \textbf{Kernel parameters} & \textbf{CIFAR-10} & \textbf{CIFAR-100} \\ 
      \specialrule{0.5pt}{2pt}{2pt}
      ResNet-20  & \numprint{272282} & \numprint{270896} & 92.46\% & -- \\
      ResNet-56  & \numprint{855578} & \numprint{851504} & 93.71\% & 71.90\% \\
      ResNet-110 & \numprint{1730522} & \numprint{1722416} & 94.29\% & 72.21\% \\
      \specialrule{0.5pt}{2pt}{2pt}
    \end{tabular}
  \end{center}
\caption{ResNets architecture description, including baseline accuracy across datasets.}
\label{tab:resnet}
\end{table}


\textbf{Hyper-parameters}

Learning rate started with $0.1$ and was multiplied by $0.1$ twice, after \numprint{36000} and \numprint{54000} iterations. One training cycle was \numprint{72000} iterations in total.
For all batch normalization layers, we set the batch norm decay to 0.997, following \cite{Renda} which was also used in the original TensorFlow implementation\footnote{\url{https://github.com/tensorflow/models/blob/r1.13.0/official/resnet/resnet_model.py}}.
We initialize network's weights with what is known as He uniform initialization from \cite{he_uniform}.
We regularize ResNets, during both training and finetuning, using $L2$ penalty with $10^{-4}$ coefficient.
In other words, the loss function (from which we calculate the gradients) looks as follows:
$$\texttt{FinalLoss} = \texttt{CategoricalCrossentropy}(\texttt{GroundTruth}, \texttt{Prediction}) + 10^{-4} \times \sum_{i\in W} w_i^2$$

\subsubsection{Wide Residual Networks (Wide ResNet, WRN)}

WRN networks were introduced in \cite{wrn}.
They are networks created by simply increasing the number of filters in preactivated ResNet networks (\cite{resnetv2}).

\begin{table}[H]
\small
\setlength{\tabcolsep}{20pt}
  \begin{center}
    \begin{tabular}{l|c|c|c}
      \specialrule{1pt}{2pt}{2pt}
      \textbf{Model} & \textbf{Trainable parameters} & \textbf{Kernel parameters} & \textbf{CIFAR-10}\\ 
      \specialrule{0.75pt}{2pt}{2pt}
      WRN-16-8 & \numprint{10961370} & \numprint{10954160} & 95.72\% \\
      \specialrule{0.75pt}{2pt}{2pt}
    \end{tabular}
  \end{center}
\caption{Wide ResNet architecture description.}
\label{tab:wrn}
\end{table}

\textbf{Hyper-parameters}

\nopagebreak
As Wide ResNets are newer and much larger than ResNets, hyper-parameters are slightly different.
To choose them, we follow \cite{wrn}. Learning rate starts with $0.1$ and multiplied by $0.2$ thrice: after \numprint{32000}, \numprint{48000} and \numprint{64000} iterations. Training lasts for \numprint{80000} iterations. For all batch normalization layers, we use hyper-parameters from the newer TensorFlow implementation\footnote{\url{https://github.com/tensorflow/models/blob/r2.5.0/official/vision/image_classification/resnet/resnet_model.py}} with batch norm decay set to 0.9. Following \cite{wrn}, we use larger $L2$ penalty for this network: $2\times10^{-4}$. Finally, the loss function is as follows:
$$\texttt{FinalLoss} = \texttt{CategoricalCrossentropy}(\texttt{GroundTruth}, \texttt{Prediction}) + 2 \times 10^{-4} \times \sum_{i\in W} w_i^2$$

\subsection{Datasets}

CIFAR-10 and CIFAR-100 are image classification datasets introduced in \cite{cifar10}. Following \cite{Renda}, we use all (\numprint{50000}) training examples to train the model.

\begin{table}[H]
\small
\setlength{\tabcolsep}{14pt}
  \begin{center}
    \begin{tabular}{l|c|c|c|c}
      \specialrule{1pt}{2pt}{2pt}
\textbf{Dataset} & \textbf{Training examples} & \textbf{Validation examples} & \textbf{Classes} & \textbf{Resolution}\\ 
      \specialrule{0.5pt}{2pt}{2pt}
      CIFAR-10  & \numprint{50000} & \numprint{10000} & 10 & 32$\times$32\\
      CIFAR-100  & \numprint{50000} & \numprint{10000} & 100 & 32$\times$32\\
      \specialrule{0.5pt}{2pt}{2pt}
    \end{tabular}
  \end{center}
\caption{CIFAR datasets description.}
\label{tab:cifar}
\end{table}

\subsubsection{Postprocessing}
We used a standard postprocessing for both CIFAR-10 and CIFAR-100 datasets (\cite{Renda}, \cite{Frankle}, \cite{wrn}). During training and just before passing data to the model, we:
\begin{enumerate}
    \item standardized the input by subtracting the mean and dividing by the std of RGB channels (calculated on training dataset),
    \item randomly flipped in horizontal axis,
    \item added a four pixel reflection padding,
    \item randomly cropped the image to its original size.
\end{enumerate}

During the validation, we did only the first step of the above.


\subsection{Experimental setup and code}

Our ready-to-use code, which includes experiment definitions, can be found at 
\url{https://github.com/gahaalt/reproducing-comparing-rewinding-and-finetuning}.
It's written using TensorFlow (\cite{tensorflow}) version 2.4.2 in Python. More details are included in the repository.

\subsection{Computational requirements}

Recreating the experiments required a modern GPU, training all models on CPU was virtually impossible. Training time varies depending on a lot of factors: network version and size, exact version of the deep learning library, and even the operating system. In our case, using TensorFlow 2.4.2 on Ubuntu and a single RTX 3080 GPU, the smallest of the used models, ResNet-20, takes about 20 minutes to train on CIFAR-10 dataset. To replicate our experiments, training at least a single baseline network and then, separately, a single pruned network, is required. To reduce computational requirements, we reused one dense baseline for multiple compression ratios. Approximated training time requirements can be seen in the table below.

\begin{table}[H]
\small
\setlength{\tabcolsep}{12pt}
  \begin{center}
    \begin{tabular}{l|c|c|c|c}
      \specialrule{1pt}{2pt}{2pt}
\textbf{Model} & \textbf{Dataset} & \textbf{Number of iterations} & \textbf{Iterations per second} & \textbf{Time for training cycle}\\ 
      \specialrule{0.5pt}{2pt}{2pt}
      ResNet-20  & CIFAR-10 & \numprint{72000} & 59.0 & 22 min \\
      ResNet-56  & CIFAR-10 & \numprint{72000} & 28.6 & 43 min \\
      ResNet-110  & CIFAR-10 & \numprint{72000} & 15.9 & 77 min \\
      WRN-16-8  & CIFAR-10 & \numprint{80000} & 17.4 & 78 min \\
      \specialrule{0.5pt}{2pt}{2pt}
    \end{tabular}
  \end{center}
\caption{Time requirements for replicating or running experiments from this report. Reported times are obtained using a single RTX 3080 GPU in Linux environment, using TensorFlow in version 2.4.2.}
\label{tab:compute}
\end{table}

For all our experiments in total, we used around 536 GPU hours.

\section{Method description}


We compare three methods of retraining after pruning. For all of them, the starting point is a network that was already trained to convergence, then pruned to a desired sparsity. The difference between the three retraining methods is what follows after it.

\subsection{Fine-tuning}
Fine-tuning is retraining with a small, constant learning rate -- in our case, whenever fine-tuning was used, the learning rate was set to 0.001 as in \cite{Renda}. We finetune the network for the same number of iterations as the baseline -- \numprint{72000} iterations in the case of the original ResNet architecture. In this method, such long retraining would not be necessary in practical applications, since the network converges much faster.

\subsection{Weight rewinding}
Weight rewinding restores the network's weights from a previous point (possibly beginning) in the training history and then continues training from this point using the original training schedule -- in our case a piecewise constant decaying learning rate schedule.
When rewinding a network to iteration $K$ that originally trained for $N$ iterations: first prune the dense network that was trained for $N$ iterations. Then, for connections that survived, restore their values to $K$-th iteration from the training history. Then train to the convergence for the remaining $N-K$ iterations.

\subsection{Learning rate rewinding}
Learning rate rewinding continues training with weights that have already converged, but restores the learning rate schedule to the beginning, just as if we were training from scratch, and then trains to the convergence once again. This reminds the cyclical learning rates from \cite{cyclical}. Learning rate rewinding really is weight rewinding for $K = N$, but the final retraining is always for $N$ iterations.

\section{Results}
In most of our experiment, just as \cite{Renda}, we investigate how does the trade-off between prediction accuracy and compression ratio look like. 
In one of the experiments (Table \ref{tab:resnet110}) we verify only one compression ratio, but for the rest, we verify multiple.
We report a median result out of 2 up to 12 trials for each compression ratio. To better utilize our compute capabilities, we decided to spend more training cycles in situations where there is no clear winner between the compared methods. On each plot, we include error bars showing 80\% confidence intervals.

\subsection{Results reproducing original paper}

In this section, we include experiments that we successfully reproduced.
They match the original ones within 1\% error margin.

Across all scenarios where finetuning was tested, it was by far the worst of the three methods, which directly supports claim 1 (Section \ref{sec:claims}). Weight rewinding and learning rate rewinding most often are equally matched, but in some cases learning rate rewinding works a little better.

\textbf{ResNets on CIFAR-10 dataset}
\nopagebreak

\begin{figure}[H]
\centering
\setlength{\tabcolsep}{0pt}
\begin{tabular}{c}
\includegraphics[width=1.0\linewidth]{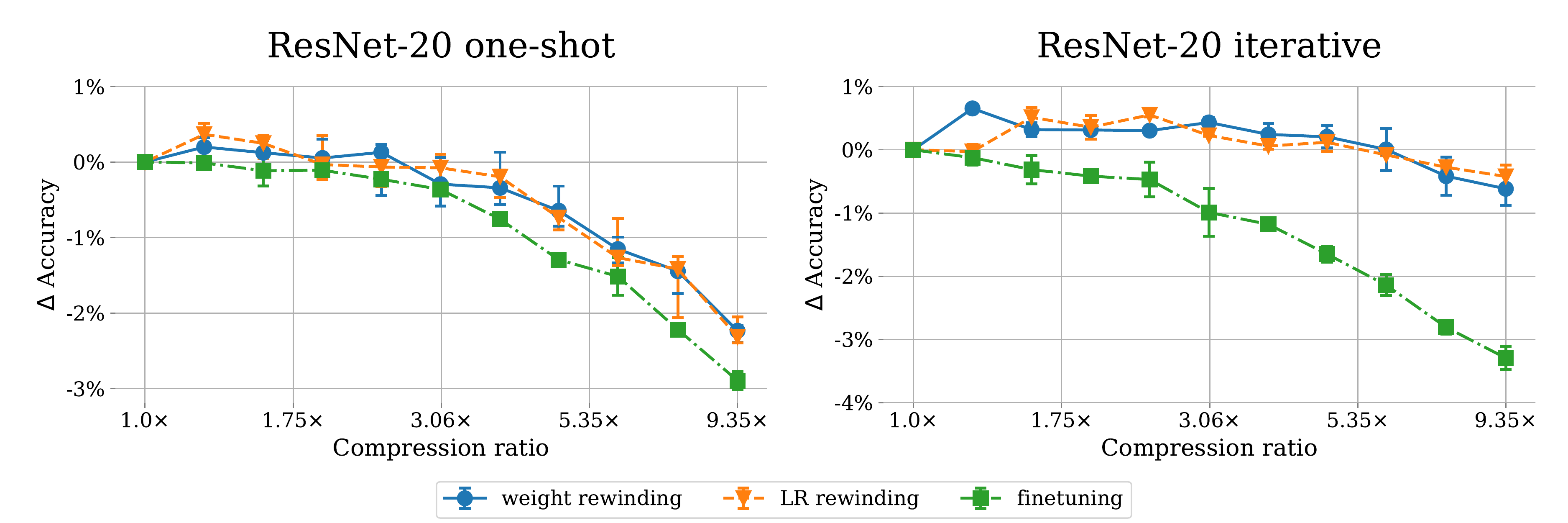}
\end{tabular}
\caption{Results of ResNet-20 (Table \ref{tab:resnet}) on CIFAR-10 (Table \ref{tab:cifar}) with unstructured, magnitude pruning in versions: one-shot and iterative. Results show varying compression ratios. Maximal compression ratio (9.35$\times$) means that there are only \numprint{29000} non-zero kernel parameters left. This experiment supports claims 1, 2, 3 (Section \ref{sec:claims}).}
\label{fig:resnet20-1}
\end{figure}

\begin{figure}[H]
\setlength{\tabcolsep}{0pt}
\centering
    \begin{tabular}{c}
      \includegraphics[width=1.0\linewidth]{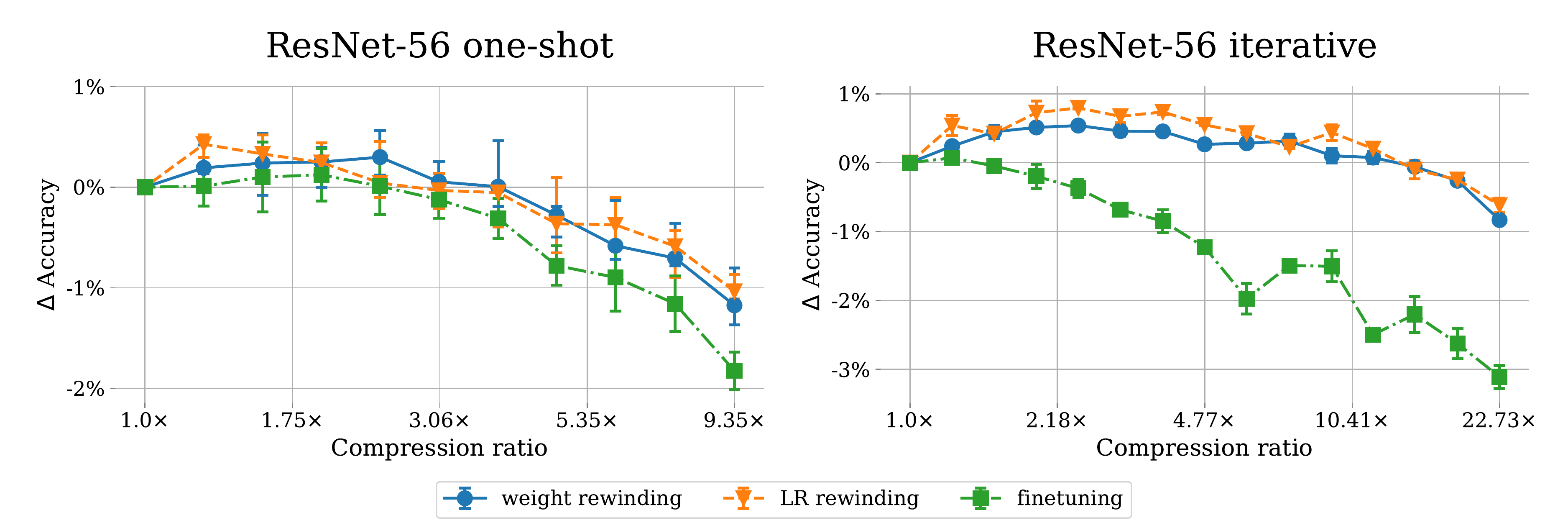}
    \end{tabular}
\caption{Results of ResNet-56 (Table \ref{tab:resnet}) on CIFAR-10 (Table \ref{tab:cifar}) with unstructured, magnitude pruning in versions: one-shot and iterative. Results with varying compression ratios. Maximal compression ratio means (22.73$\times$) that there are only \numprint{37600} non-zero kernel parameters left. This experiment supports claims 1, 2, 3 (Section \ref{sec:claims}).}
\label{fig:resnet56}
\end{figure}

\begin{table}[H]
\small
\setlength{\tabcolsep}{14pt}
  \begin{center}
    \begin{tabular}{l|c|c|c|c}
      \specialrule{1pt}{2pt}{2pt}
\textbf{Network} & \textbf{Dataset} & \textbf{Retraining} & \textbf{Sparsity} & \textbf{Test Accuracy} \\ 
      \specialrule{0.75pt}{2pt}{2pt}
      ResNet-110  & CIFAR-10 & None & 0\% & 94.29\% \\
      ResNet-110  & CIFAR-10 & LR rewinding & 89.3\% & 93.74\% \\
      ResNet-110  & CIFAR-10 & weight rewinding & 89.3\% & 93.73\% \\
      ResNet-110  & CIFAR-10 & finetuning & 89.3\% & 93.32\% \\
      \specialrule{0.75pt}{2pt}{2pt}
    \end{tabular}
  \end{center}
\caption{Results of ResNet-110 (Table \ref{tab:resnet}) trained on CIFAR-10 (Table \ref{tab:cifar}) with unstructured, one-shot magnitude pruning. Sparsity $89.3\%$ corresponds to $9.35\times$ compression ratio. This experiment supports claims 1, 2, 3 (Section \ref{sec:claims}).}
\label{tab:resnet110}
\end{table}

\begin{figure}[H]
\setlength{\tabcolsep}{0pt}
\centering
    \begin{tabular}{c}
      \includegraphics[width=0.7\linewidth]{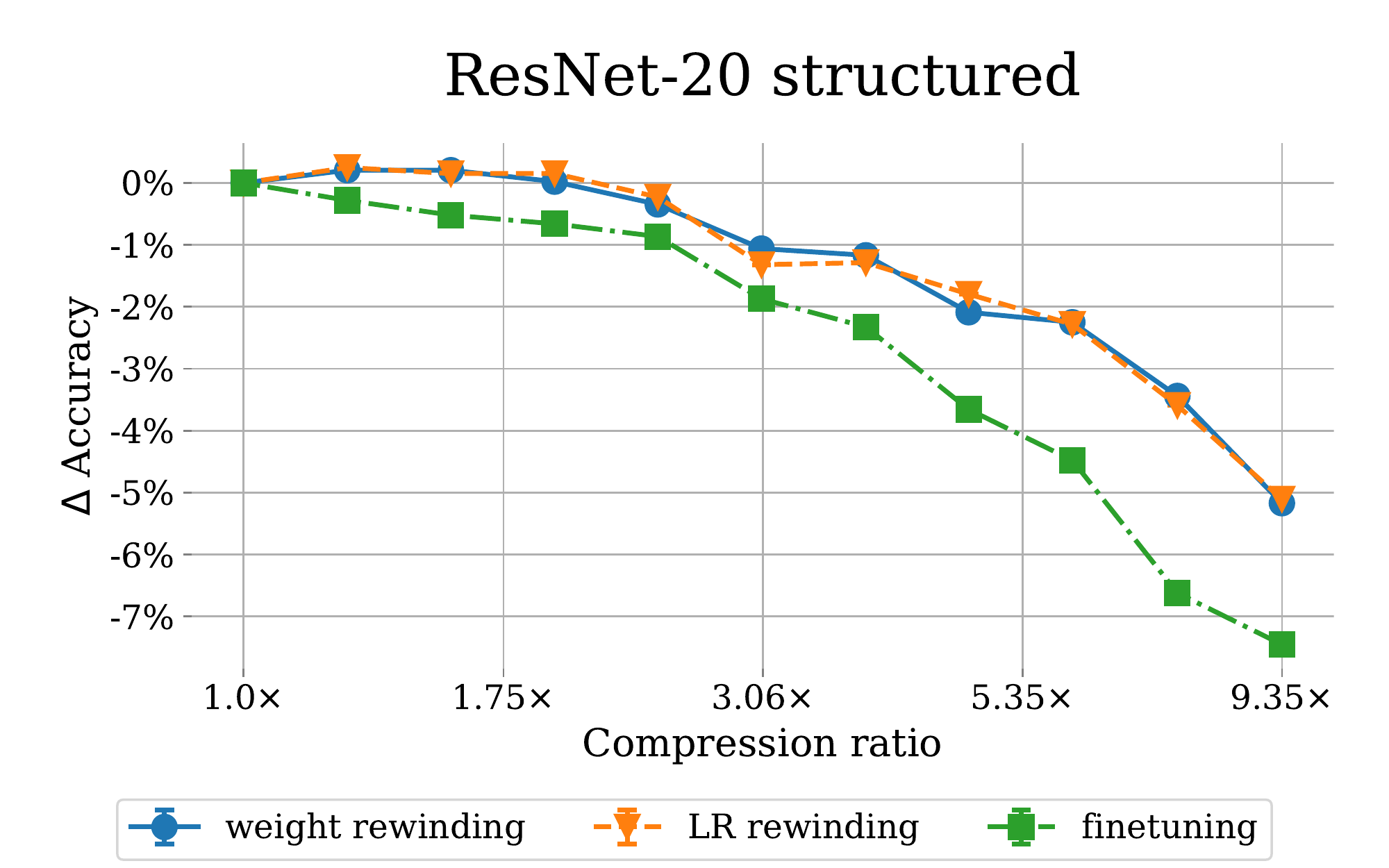}
    \end{tabular}
\caption{Results of ResNet-20 (Table \ref{tab:resnet}) on CIFAR-10 (Table \ref{tab:cifar}) with structured, one-shot, magnitude pruning. Results show varying compression ratios. Maximal compression ratio (9.35$\times$) means that there are only \numprint{29000} non-zero kernel parameters left in ResNet-20.}
\label{fig:resnet20-2}
\end{figure}



\subsection{Results beyond original paper}
 
\textbf{ResNets on CIFAR-100 dataset}
\nopagebreak
\begin{figure}[H]
\setlength{\tabcolsep}{0pt}
\centering
    \begin{tabular}{c}
      \includegraphics[width=0.7\linewidth]{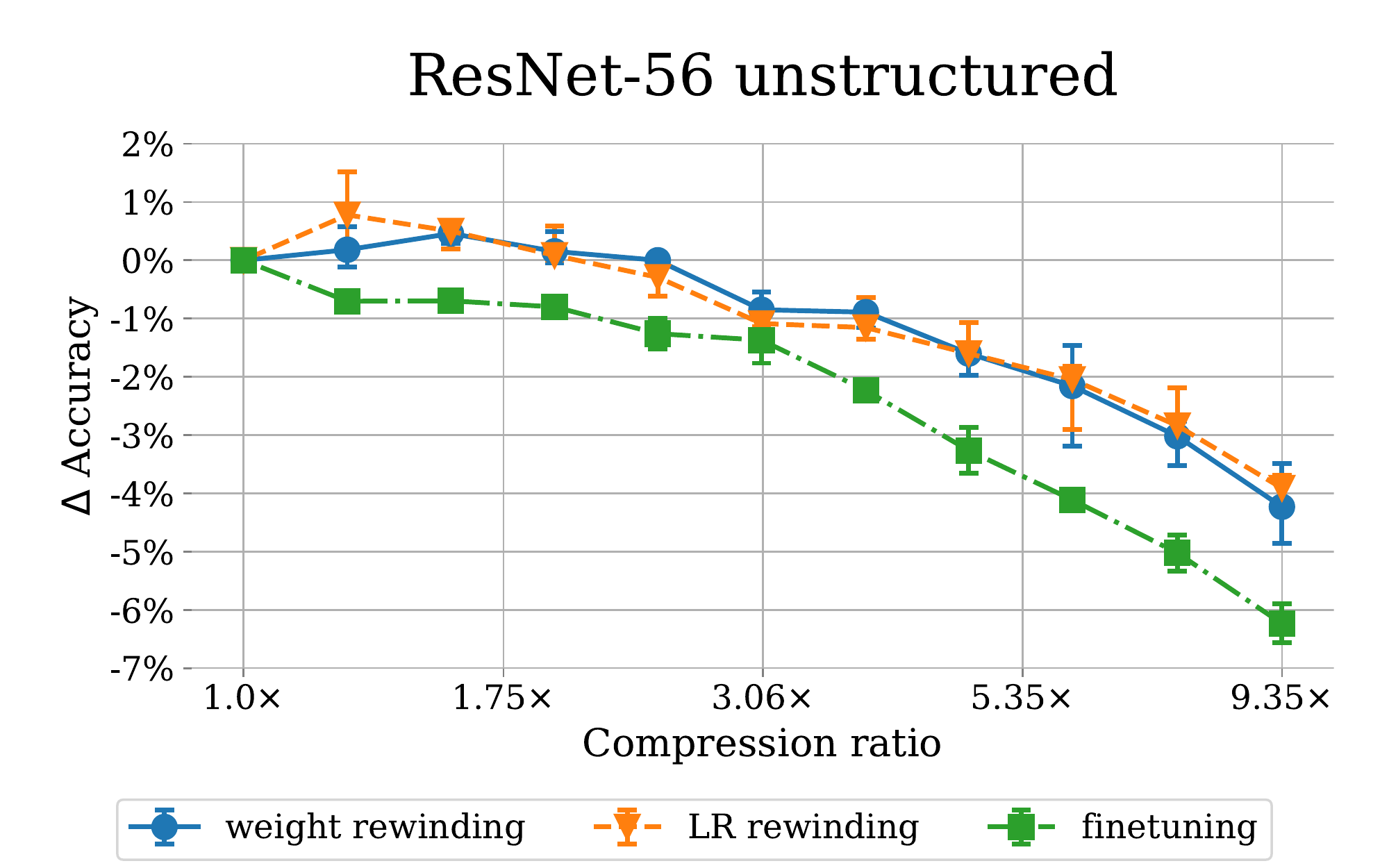}
    \end{tabular}
\caption{Results of ResNet-56 (Table \ref{tab:resnet}) on CIFAR-100 (Table \ref{tab:cifar}) with unstructured, one-shot, magnitude pruning. Results with varying compression ratios. Maximal compression ratio (9.35$\times$) means that there are only \numprint{91500} non-zero kernel parameters left. This experiment supports claims 1, 2, 3 (Section \ref{sec:claims}) even though this scenario wasn't originally tested in \cite{Renda}.}
\label{fig:resnet56-2}
\end{figure}

\textbf{WRN-16-8 on CIFAR-10 dataset}
\nopagebreak

\begin{figure}[H]
\setlength{\tabcolsep}{0pt}
  \begin{center}
    \begin{tabular}{c}
      \includegraphics[width=1.0\linewidth]{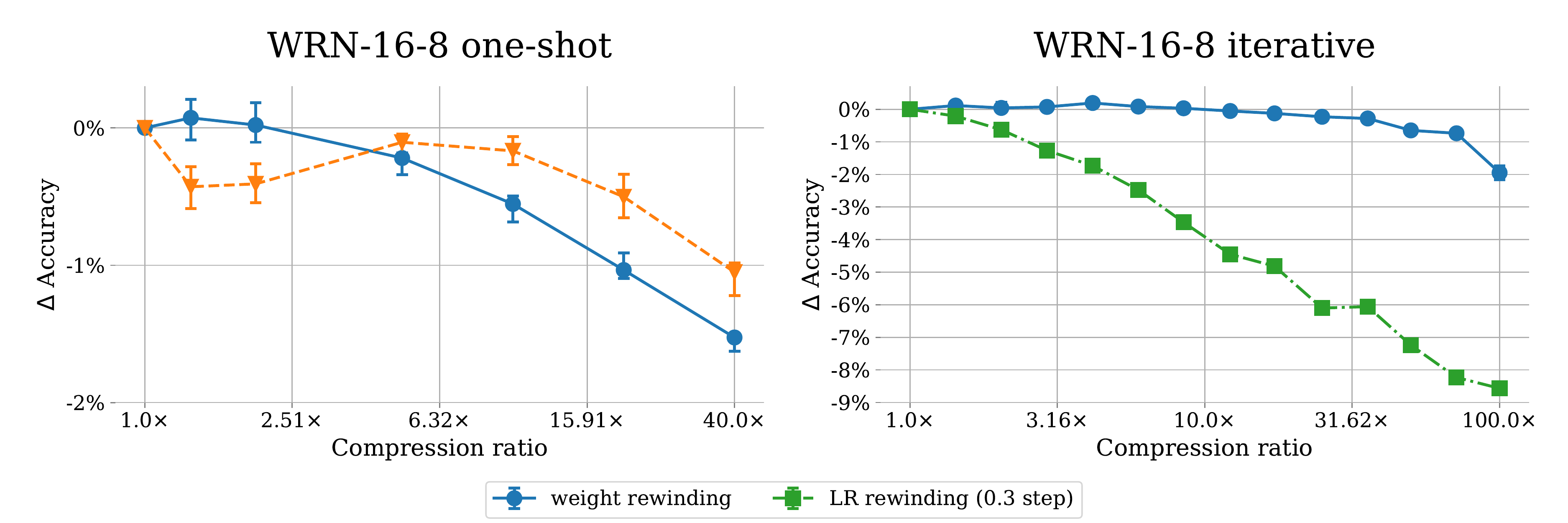}\\
    \end{tabular}
  \end{center}
\caption{Results of WRN-16-8 (Table \ref{tab:wrn}) on CIFAR-10 (Table \ref{tab:cifar}) with unstructured, magnitude pruning in versions: one-shot and iterative. Results with varying compression ratios. Maximal compression ratio (100$\times$) leaves \numprint{109500} non-zero kernel parameters while achieving around 94\% accuracy or around 95\% when leaving \numprint{153400} non-zero parameters. One can see catastrophic effects of low-sparsity pruning when using learning rate rewinding procedure.}
\label{fig:wrn-1}
\end{figure}

WRN-16-8 shows consistent behaviour -- accuracy in the low sparsity regime is reduced in comparison to the baseline. In the case of iterative pruning, where each step is another pruning in the low sparsity regime, it leads to a large difference between the two retraining methods. Since for WRN-16-8 one-shot, low sparsity pruning shows a small regression in comparison to the baseline, this regression accumulates when pruning multiple times, as we do in iterative pruning. This can be seen in Figure \ref{fig:wrn-1}.

For iterative pruning (figures \ref{fig:resnet20-1}, \ref{fig:resnet56}) we used a nonstandard step size of 30\% per iterative pruning iteration, which was a way to reduce the computational requirements. We provide a comparison of our step size to the more commonly used 20\%. We show that there is virtually no difference between both versions and the aforementioned catastrophic degradation occurs in both cases, as long as the step size is in the low sparsity regime.

\begin{figure}[H]
\setlength{\tabcolsep}{0pt}
  \begin{center}
    \begin{tabular}{c}
      \includegraphics[width=0.7\linewidth]{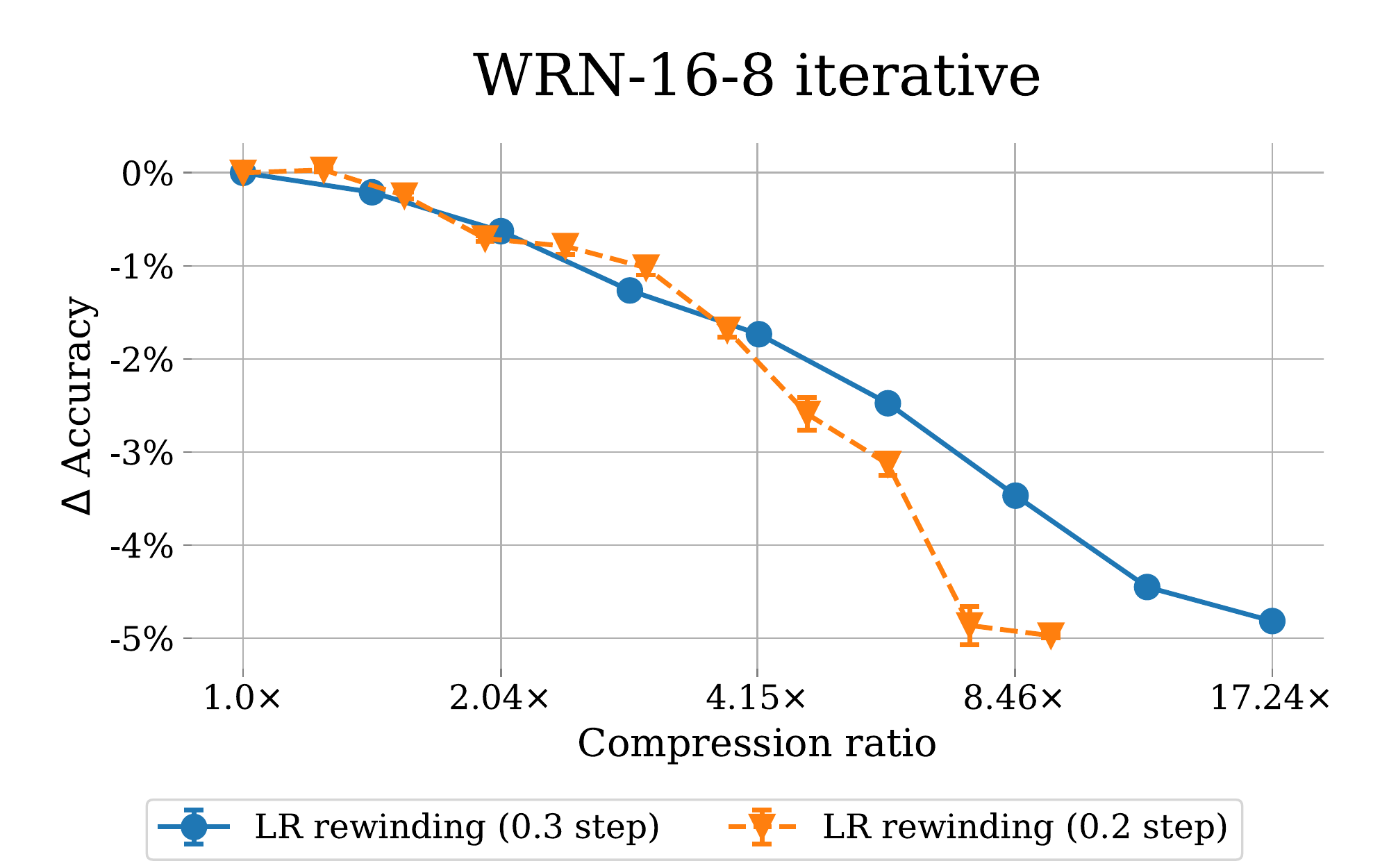}\\
    \end{tabular}
  \end{center}
\caption{Results of WRN-16-8 (Table \ref{tab:wrn}) on CIFAR-10 (Table \ref{tab:cifar}) with unstructured, iterative, magnitude pruning with two different step sizes. Results show varying compression ratios and accuracy.}
\label{fig:wrn-2}
\end{figure}

\section{Discussion}

We were able to confirm the general conclusion of \cite{Renda}. Fine-tuning can mostly be replaced by other retraining techniques, e.g., by weight rewinding as done by \cite{Frankle}. However, we have also shown in Figure \ref{fig:wrn-1} that the newly proposed learning rate rewinding is a poor choice when we are pruning larger networks -- in our case that is WRN-16-8. We believe this should be further examined as there might exist a simple workaround to this problem -- a retraining procedure in between weight rewinding and learning rate rewinding, which works in all cases. Furthermore, it would be interesting to see where exactly learning rate rewinding starts losing accuracy in comparison to weight rewinding and why this catastrophic accuracy degradation occurs. Perhaps, the reason for it not occurring with the original ResNet architecture is the degree to which the larger networks overtrain -- larger networks tend to overtrain more. Such an overtrained network might not be a good starting point for the retraining.


\section*{Acknowledgements}
The authors thank Polish National Science Center for funding
under the OPUS-18 2019/35/B/ST6/04379 grant and the PlGrid
consortium for computational resources.

\bibliography{bib}

\end{document}